\documentclass[conference]{IEEEtran}
\ifCLASSOPTIONcompsoc
\usepackage[nocompress]{cite}
\else
\usepackage{cite}
\fi

\makeatletter
\let\@ORGmakecaption\@makecaption
\long\def\@makecaption#1#2{\@ORGmakecaption{#1}{#2}\vskip\belowcaptionskip\relax}
\makeatother

\usepackage{upquote}
\usepackage{url, listings, float}
\usepackage{graphicx, subfigure, makecell}

\usepackage{booktabs} 
\usepackage{amsmath}
\usepackage{multirow}

\usepackage[utf8]{inputenc}

\def\ei{EI\xspace}
\def\ec{EC\xspace}
\def\ai{AI\xspace}
\def\oei{\textit{OpenEI}\xspace}
\def\libei{{\tt libei}\xspace}
\def\packagemanager{\textit{package manager}\xspace}
\def\modelselector{\textit{model selector}\xspace}

\def\etal{\textit{et al.}\xspace}
\def\etc{\textit{etc.}\xspace}

\usepackage{verbatim}
\usepackage{color}
\usepackage{xcolor}
\usepackage{xspace}
\usepackage{ifthen}
\usepackage{url}
\newboolean{showcomments}
\setboolean{showcomments}{true}

\newcommand{\rev}[1]{{\color{black}{}#1}}

\newcommand{\todo}[1]{\ifthenelse{\boolean{showcomments}}
	{ \textcolor{blue}{(To do:  #1)}}{}}
\newcommand{\note}[1]{\ifthenelse{\boolean{showcomments}}
	{ \textcolor{blue}{(Cmt:  #1)}}{}}
\newcommand{\toall}[1]{\ifthenelse{\boolean{showcomments}}
    {\textcolor{red}{To All: #1}}{}}

\colorlet{punct}{red!60!black}
\definecolor{background}{HTML}{EEEEEE}
\definecolor{delim}{RGB}{20,105,176}
\colorlet{numb}{magenta!60!black}

\begin{document}

\title{OpenEI: An Open Framework for Edge Intelligence}

\author{
\IEEEauthorblockN{
Xingzhou~Zhang\IEEEauthorrefmark{1}\IEEEauthorrefmark{2},
Yifan~Wang\IEEEauthorrefmark{1}\IEEEauthorrefmark{2},
Sidi~Lu\IEEEauthorrefmark{1},
Liangkai~Liu\IEEEauthorrefmark{1},
Lanyu~Xu\IEEEauthorrefmark{1} and
Weisong~Shi\IEEEauthorrefmark{1}
}

\IEEEauthorblockA{
\IEEEauthorrefmark{1}Department of Computer Science, Wayne State University, Detroit, MI, USA, 48202\\
\IEEEauthorrefmark{2}Institute of Computing Technology, University of Chinese Academy of Sciences, Beijing, China, 100190\\
\{zhangxingzhou, wangyifan2014\}@ict.ac.cn, \{lu.sidi, liangkai, xu.lanyu, weisong\}@wayne.edu
}
}

\maketitle
	\footnote{This paper has been accepted by ICDCS 2019. Please cite: Xingzhou Zhang, Yifan Wang, Sidi Lu, Liangkai Liu, Lanyu Xu, Weisong Shi, \textit{OpenEI: An Open Framework for Edge Intelligence}, in Proceedings of the 39th IEEE International Conference on Distributed Computing Systems (ICDCS), July 7-10, 2019, Dallas, USA.}
\begin{abstract}

In the last five years, edge computing has attracted tremendous attention from industry and academia due to its promise to reduce latency, save bandwidth, improve availability, and protect data privacy to keep data secure. At the same time, we have witnessed the proliferation of AI algorithms and models which accelerate the successful deployment of intelligence mainly in cloud services. These two trends, combined together, have created a new horizon: Edge Intelligence (EI). The development of EI requires much attention from both the computer systems research community and the AI community to meet these demands.

However, existing computing techniques used in the cloud are not applicable to edge computing directly due to the diversity of computing sources and the distribution of data sources. We envision that there missing a framework that can be rapidly deployed on edge and enable edge AI capabilities. To address this challenge, in this paper we first present the definition and a systematic review of EI. Then, we introduce an Open Framework for Edge Intelligence (OpenEI), which is a lightweight software platform to equip edges with intelligent processing and data sharing capability. We analyze four fundamental EI techniques which are used to build OpenEI and identify several open problems based on potential research directions. Finally, four typical application scenarios enabled by OpenEI are presented.




\end{abstract}


\IEEEpeerreviewmaketitle
\section{Introduction}



With the burgeoning growth of the Internet of Everything, the amount of data generated by edge increases dramatically, resulting in higher network bandwidth requirements. 
Meanwhile the emergence of novel applications calls for lower latency of the network. Based on these two main requirements, \ec arises, which refers to processing the data at the edge of the network.
Edge Computing (\ec) guarantees quality of service when dealing with a massive amount of data for cloud computing \cite{shi2016edge}. Cisco Global Cloud Index \cite{Cisco18} estimates that there will be 10 times more useful data being created (85 ZB) than being stored or used (7.2 ZB) by 2021, and \ec is a potential technology to help bridge this gap. 

\begin{figure}[!htp]
	\centering
	\includegraphics[width=0.85\linewidth]{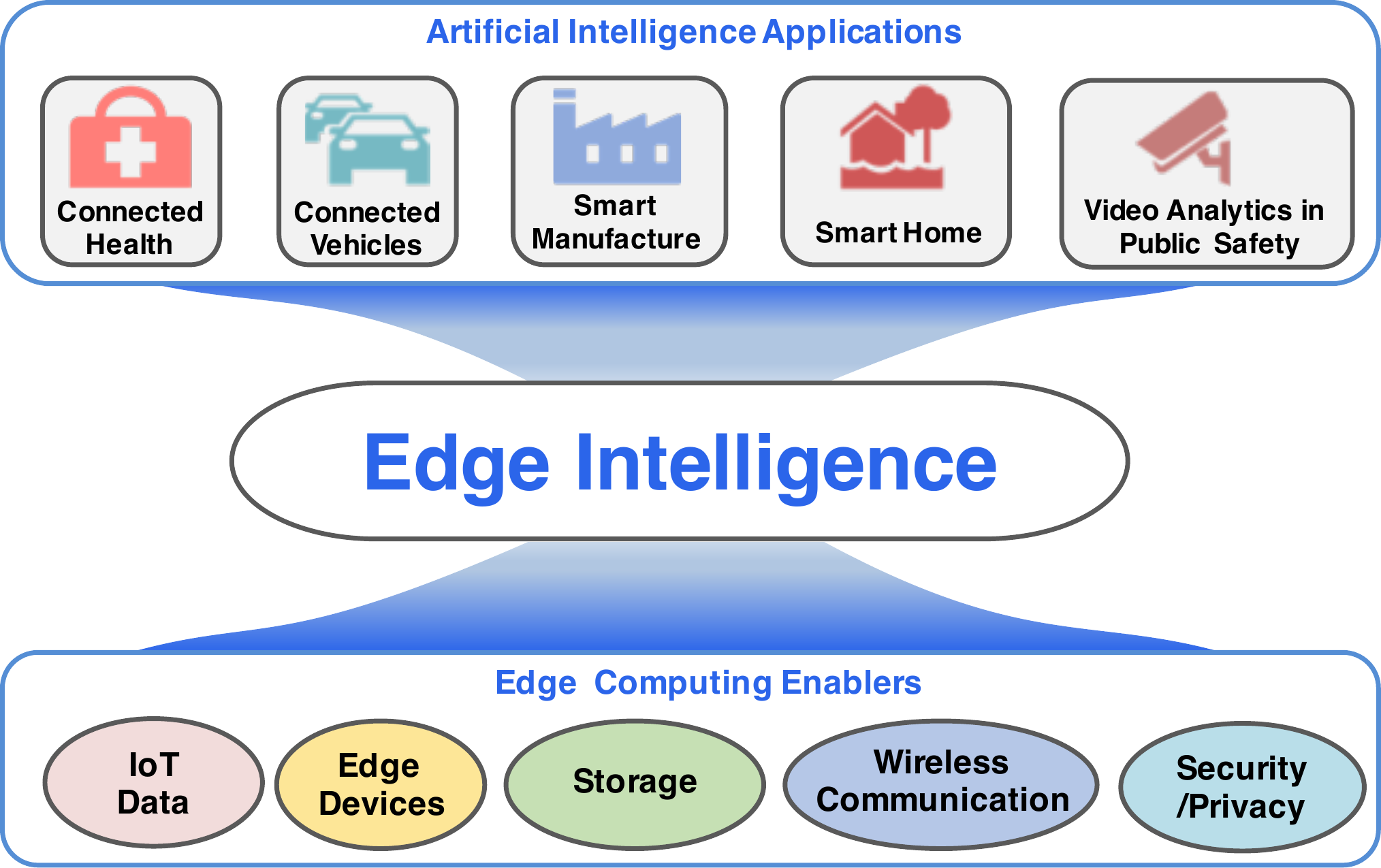}
	\caption{Motivation of Edge Intelligence.}
	\label{fig:Motivation of Edge Intelligence.}
\end{figure}

At the same time, Artificial Intelligence (\ai) applications based on machine learning (especially deep learning algorithms) are fueled by advances in models, processing power, and big data. 
Nowadays, applications are built as a central attribute, and users are beginning to expect near-human interaction with the appliances they use.
For example, since the sensors and cameras mounted on an autonomous vehicle generate about one gigabyte of data per second \cite{Self-driving_cars_could_create_1GB_data_a_second}, it is hard to upload the data and get instructions from the cloud in real-time. As for mobile phone applications, such as those related with face recognition and speech translation, they have high requirements for running either online or offline.

Pushed by \ec techniques and pulled by \ai applications, Edge Intelligence (\ei) has been pushed to the horizon. As is shown in Figure \ref{fig:Motivation of Edge Intelligence.}, the development of \ec techniques, including powerful IoT data, edge devices, storage, wireless communication, and security and privacy make it possible to run \ai algorithms on the edge. \ai applications, including connected health, connected vehicles, smart manufacturing, smart home, and video analytics call for running on the edge. In the \ei scenario, advanced AI models based on machine learning algorithms will be optimized to run on the edge. The edge will be capable of dealing with video frames, natural speech information, time-series data and unstructured data generated by cameras, microphones, and other sensors without uploading data to the cloud and waiting for the response.

Migrating the \ai functions from the cloud to the edge is highly regarded by industry and academy. Forbes listed the convergence of IoT and AI on the edge as one of five \ai trends in 2019 \cite{5-artificial-intelligence-trends}. Forbes believes that most of the models trained in the public cloud will be deployed on the edge and edge devices will be equipped with special AI chips based on FPGAs and ASICs. Microsoft provides Azure IoT Edge \cite{MSedge}, a fully managed service, to deliver cloud intelligence locally by deploying and running \ai algorithms and services on cross-platform edge devices. 
Similar to Azure IoT Edge, Cloud IoT Edge \cite{Cloud_IoT_Edge} extends Google Cloud's data processing and machine learning to billions of edge devices by taking advantage of Google \ai products, such as TensorFlow Lite and Edge TPU. AWS IoT Greengrass \cite{Amazonedge} has been published to make it easy to perform machine learning inference locally on devices, using models that have been trained and optimized in the cloud. 

However, several challenges when offloading state-of-the-art \ai techniques on the edge directly, including 

\begin{itemize}
    \item \textit{Computing power limitation}. The edge is usually resource-constrained compared to the cloud data center, which is not a good fit for executing DNN represented \ai algorithms since DNN requires a large footprint on both storage (as big as 500MB for VGG-16 Model \cite{simonyan2014very}) and computing power (as high as 15300 MMA for executing VGG-16 model \cite{howard2017mobilenets}). 
    \item \textit{Data sharing and collaborating}. The data on the cloud data center is easy to be batch processed and managed, which is beneficial in terms of the concentration of data. However, the temporal-spatial diversity of edge data creates obstacles for the data sharing and collaborating.
    \item \textit{Mismatch between edge platform and \ai algorithms}. The computing power on the cloud is relatively consistent while edges have diverse computing powers. Meanwhile, different \ai algorithms have different computing power requirements. Therefore, it is a big challenge to match an existing algorithm with the edge platform.
\end{itemize}

\rev{
To address these challenges, this paper proposes an Open Framework for Edge Intelligence, \oei, which is a lightweight software platform to equip the edge with intelligent processing and data sharing capability.
To solve the problems that the \ec power limitation brings, \oei contains a lightweight deep learning package (\packagemanager) which is designed for the resource constrained edge and includes optimized \ai models. 
In order to handle the data sharing problem, \libei is designed to provide a uniform RESTful API. By calling the API, developers are able to access all data, algorithms, and computing resources. The heterogeneity of the architecture is transparent to the user, which makes it possible to share data and collaborate between edges.
In order to solve the mismatch problem, \oei designs a model selector to find the most suitable models for a specific targeting edge platform. The model selector refers to the computing power (such as memory and energy) that the algorithm requires and the edge platform provides.
}
The contributions of this paper are as follows:
\begin{itemize}
    \item A formal definition and a systematic analysis of \ei are presented. Each \ei algorithm is defined as a four-element tuple \textit{ALEM} $<Accuracy, Latency, Energy, Memory\ footprint>$. 
    \item \oei, an Open Framework for Edge Intelligence, is proposed to address the challenges of \ei, including computing power limitations, data sharing and collaborating, and the mismatch between edge platform and \ai algorithms.
    \item Four key enabling techniques of \ei and their potential directions are depicted. Several open problems are also identified in the paper. 
\end{itemize}


The remainder of this paper is organized into six sections. 
In Section \ref{sec: The definition of Edge Intelligence}, we define \ei and list the advantages of \ei. We present \oei to support \ei in Section \ref{sec: OpenEI}. Four key techniques that enable \ei are explained in Section \ref{sec: Key techniques}, including algorithms, packages, running environments, and hardware. \ei is designed to support many potential applications, such as live video analytic for public safety, connected and autonomous driving, smart home, and smart and connected health, which are illustrated in Section \ref{sec: Potential applications}. Finally, Section \ref{sec: Conclusion} concludes the paper.

\section{The definition of Edge Intelligence}
\label{sec: The definition of Edge Intelligence}

\begin{figure}[h]
	\centering
	\includegraphics[width=0.96\linewidth]{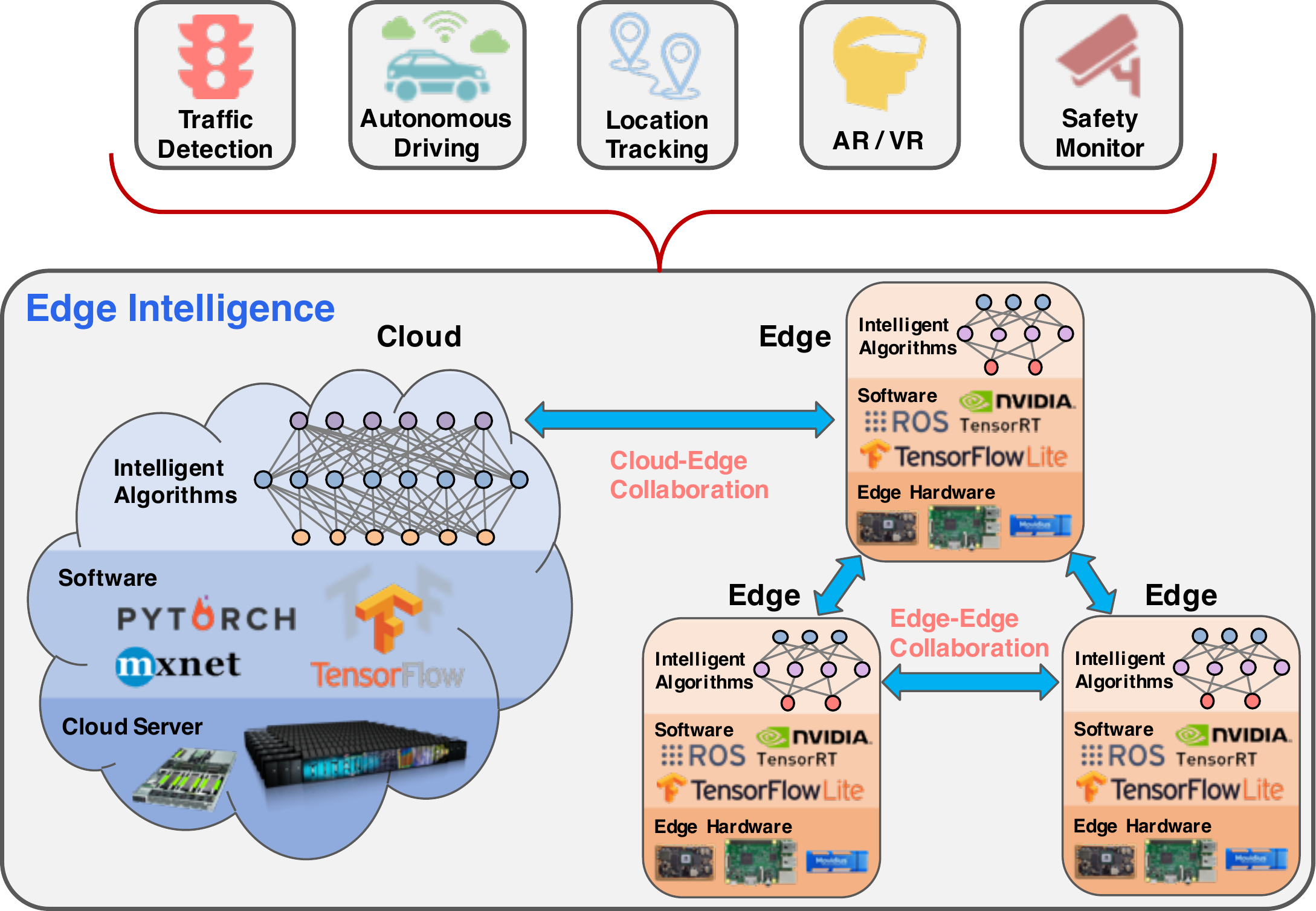}
	\caption{Edge Intelligence.}
	\label{fig:Edge intelligence}
\end{figure}

\subsection{Motivation}

The development of \ei comes from two aspects. 
On the one hand, the burgeoning growth of the IoT results in a dramatically increasing amount of IoT data, which needs to be processed on the edge. In this paper, \textbf{we define IoT as the billions of physical devices around the world that are securely connected to the Internet, individually or collaboratively, collecting and sharing data, applying intelligence to actuate the physical world in a safe way.} 
On the other hand, the emergence of \ai applications calls for a higher requirement for real-time performance, such as autonomous driving, real-time translation, and video surveillance. 
\ei is presented to deal with this massive edge data in an intelligent manner.


\subsection{Definition}
Currently, many studies related to \ei are beginning to emerge.
International Electrotechnical Commission (IEC) defines \ei as the process of when the data is acquired, stored and processed with machine learning algorithms at the network edge. It believes that several information technology and operational technology industries are moving closer to the edge of the network so that aspects such as real-time networks, security capabilities, and personalized/customized connectivity are addressed \cite{edgeintelligence_iec}. 
In 2018, \cite{plastiras2018edge} discussed the challenges and the opportunities that \ei created by presenting a use-case showing that the careful design of the convolutional neural networks (CNN) for object detection would lead to real-time performance on embedded edge devices.  \cite{zhang2018enabling} enabled \ei for activity recognition in smart homes from multiple perspectives, including architecture, algorithm and system. 

In this paper, \textbf{we define \ei as the capability to enable edges to execute artificial intelligence algorithms.} The diversity of edge hardware results in different in \ai models or algorithms they carry; that is, edges have different \ei capabilities. 
Here the capability is defined as a four-element tuple $<Accuracy, Latency, Energy, Memory\ footprint>$ which is abbreviated as \textit{ALEM}.
\textit{Accuracy} is the internal attribute of \ai algorithms. In practice, the definition of \textit{Accuracy} depends on specific applications; for example, it is measured by the mean average precision (mAP) in object detection tasks and measured by the BLEU score metric in machine translation tasks. To execute the \ai tasks on the edge, some algorithms are optimized by compressing the size of the model, quantizing the weight and other methods that will decrease accuracy. 
Better \ei capability means that the edge is able to employ the algorithms with greater \textit{Accuracy}.
\textit{Latency} represents the inference time when running the trained model on the edge. To measure the \textit{Latency}, we calculate the average latency of multiple inference tasks. When running the same models, the \textit{Latency} measures the level of performance of the edge.
\textit{Energy} refers to the increased power consumption of the hardware when executing the inference task.
\textit{Memory footprint} is the memory usage when running the \ai model.
\textit{Energy} and \textit{Memory footprint} indicate the computing resource requirements of the algorithms.

\ei involves much knowledge and technology, such as AI algorithm design, software and system, computing architecture, sensor network and so on. Figure \ref{fig:Edge intelligence} shows the overview of \ei. To support \ei, many techniques have been developed, called \ei techniques, which include algorithms, software, and hardware. 
There is a one-to-one correspondence between the cloud and the single edge. From algorithms perspective, the cloud data centers train powerful models and the edge does the inference. With the development of \ei, the edge will also undertake some local training tasks. From the software perspective, the cloud runs the cluster operating system and deep learning framework, such as TensorFlow \cite{abadi2016tensorflow} and MXNet \cite{chen2015mxnet}. On the edge, both the embedded operating system and the stand-alone operating system are widely used. The lightweight deep learning package is used to speed up the execution, such as TensorFlow Lite \cite{TensorFlow-Lite} and CoreML \cite{CoreML}. From the hardware perspective, cloud data centers are deployed on high-performance platforms, such as GPU, CPU, FPGA, and ASIC clusters while the hardware of the edge are heterogeneous edges, such as edge server, mobile phone, Raspberry Pi, \etc

\subsection{Collaboration}
As shown in Figure \ref{fig:Edge intelligence}, there are two types of collaboration for \ei: cloud-edge and edge-edge collaboration. In the cloud-edge scenario, the models are usually trained on the cloud and then downloaded to the edge which executes the inference task. Sometimes, edges will retrain the model by transfer learning based on the data they generated. The retrained models will be uploaded to the cloud and combined into a general and global model. In addition, researchers have also focused on the distributed deep learning models over the cloud and edge. For example, DDNN~\cite{teerapittayanon2017distributed} is a distributed deep neural network architecture across the cloud and edge. 

Edge-edge collaboration has two aspects. First, multiple edges work collaboratively to accomplish a compute-intensive task. For example, several edges will be distributed when training a huge deep learning network. The task will be allocated according to the computing power. 
Second, multiple edges work together to accomplish a task with different divisions based on different environments. For example, in smart home environments, a smartphone predicts when a user is approaching home, triggering and the smart thermostat will be triggered to set the suitable temperature for the users. Individually, every task is particularly difficult, but the coordination within the edge makes it easy.

\begin{figure}[!htp]
	\centering
	\includegraphics[width=0.9\linewidth]{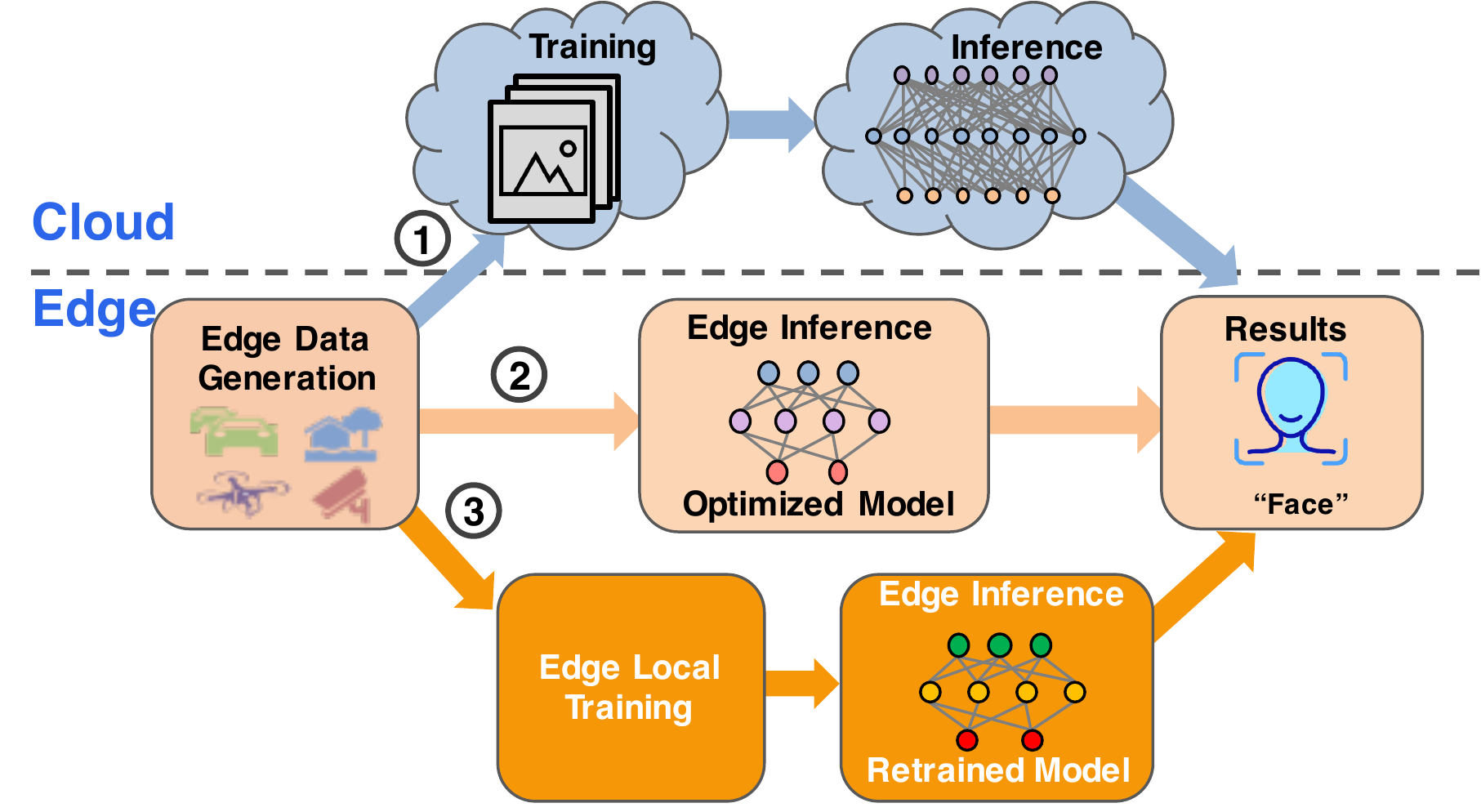}
	\caption{Dataflow of Edge Intelligence.}
	\label{fig:Dataflow of Edge Intelligence.}
\end{figure}

\subsection{Dataflow of \ei}
As is shown in Figure \ref{fig:Dataflow of Edge Intelligence.}, the data generated by the edge comes from different sources, such as cars, drones, smart homes, \etc and has three data flows:
\begin{itemize}
    \item First is uploading the data to the cloud and training based on the multi-source data. When the model training is completed, the cloud will do the inference based on the edge data and send the result to the edge. This dataflow is widely used in traditional machine intelligence.
    \item Second is executing the inference on the edge directly. The data generated by the edge will be the input of the edge model downloaded from the cloud. The edge will do the inference based on the input and output the results. This is the current \ei dataflow.
    \item Third is training on the edge locally. The data will be used to retrain the model on the edge by taking advantage of transfer learning. After retraining, the edge will build a personalized model which has better performance for the data generated on the edge. This will be the future dataflow of \ei. 
\end{itemize}

\section{\oei: Open Framework for Edge Intelligence}
\label{sec: OpenEI}
\rev{In this section, we introduce an Open Framework for Edge Intelligence (\oei), a lightweight software platform to equip the edge with intelligent processing and data sharing capability. The goal of \oei is that any hardware, ranging from Raspberry Pi to a powerful Cluster, will become an intelligent edge after deploying \oei. Meanwhile, the \ei attributes, accuracy, latency, energy, and memory footprint, will have an order of magnitude improvement comparing to the current \ai algorithms running on the deep learning package.}

\subsection{Requirements}
Let us use an example of building an \ei application to walk through the requirements of \oei. If we want to enable a new Raspberry Pi \ei capability, deploying and configuring \oei is enough. After that, the Raspberry Pi is able to detect multiple objects directly based on the data collected by the camera on board and meet the real-time requirement. It also has the ability to execute the trajectory tracking task collaborated with other \oei deployed edges. Several questions may arise: how does Raspberry Pi collect, save, and share data? How does Raspberry Pi run a powerful object detection algorithm in the real-time manner? How does Raspberry Pi collaborate with others? 

To realize the example above, \oei should meet the following four requirements: ease of use, optimal selection, interoperability, and optimization for the edge. The detailed explanations are as follows:

\textit{Ease of use}. Today, it is not very straightforward to deploy the deep learning framework and run \ai models on the edge because of the current complicated process to deploy and configure. Drawing on the idea of plug and play, \oei is deploy and play. By leveraging the API, \oei is easy to install and easy to develop third-party applications for users.

\textit{Optimal selection}. The biggest problem is not the lack of algorithms, but how to choose a matched algorithm for a specific configuration of the edge. The model selector is designed to meet the requirements.

\textit{Interoperability}. To collaborate with the cloud and other heterogeneous edges, \oei is designed as a cross-platform software. \libei provides RESTful API for the edge to communicate and work with others.

\textit{Optimization for the edge}. To run heavy \ai algorithms on the edge, being lightweight is the core feature as well as a significant difference between \oei and other data analyze platforms. Two methods are used to optimize the algorithm for the edge. One is adopting the \packagemanager which has been optimized for the edge and cutting out the redundancy operations unrelated to deep learning. The other is running lightweight algorithms which have been co-optimized with the package.

\begin{figure}[!htp]
	\centering
	\includegraphics[width=0.98\linewidth]{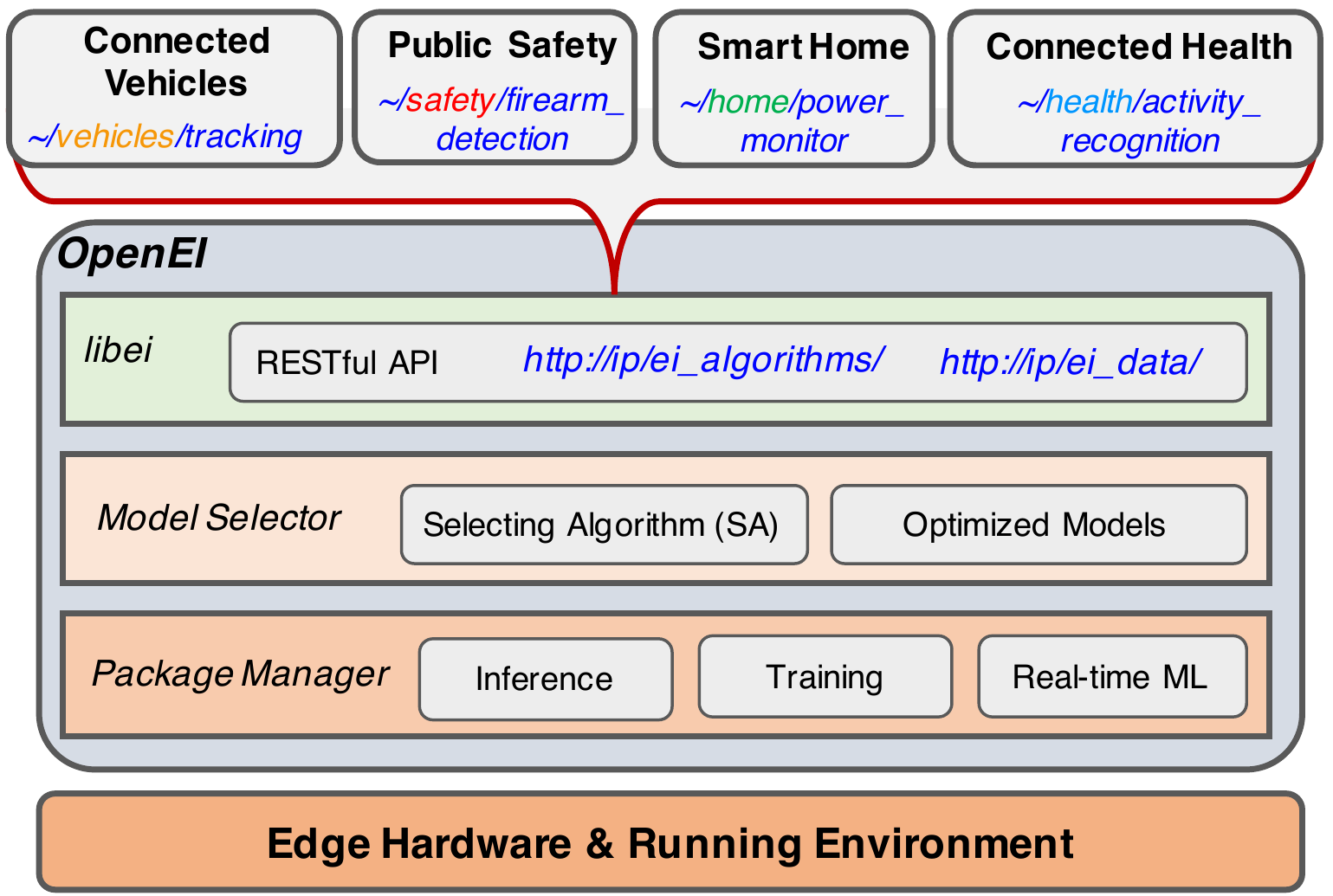}
	\caption{The overview of \oei.}
	\label{fig:The overview of OpenEI.}
\end{figure}

The answers will be found in the design of \oei. Figure \ref{fig:The overview of OpenEI.} shows the overview of \oei, which consists of three components: a \packagemanager to run inference and train the model locally, a \modelselector to select the most suitable model for the edge, and \libei, a library including a RESTful API for data sharing.

\subsection{Package Manager}
Similar to TensorFlow Lite \cite{TensorFlow-Lite}, \packagemanager is a lightweight deep learning package which has been optimized to run \ai algorithms on the edge platform, which guarantees the low power consumption and low memory footprint. \packagemanager is installed on the operating system of edge hardware and provides a running environment for \ai algorithms. In addition to supporting the inference task as TensorFlow Lite does, \packagemanager also supports training the model locally. By retraining the model based on the local data, \oei provides a personalized model which performs better than general models. 

\rev{Emerging computing challenges require real-time learning, prediction, and automated decision-making in diverse \ei domains such as autonomous vehicles and health-care informatics. To meet the real-time requirement, \packagemanager contains a real-time machine learning module. When the module is called, the machine learning task will be set to the highest priority to ensure that it has as many computing resources as possible. Meanwhile, the models are optimized for the \packagemanager since the co-optimization of the framework and algorithms is capable of increasing the system performance and speedup the execution. That is why Raspberry Pi has the ability to run a powerful object detection algorithm smoothly.}

\subsection{Model Selector}
\begin{figure}[!htp]
	\centering
	\includegraphics[width=2.5in]{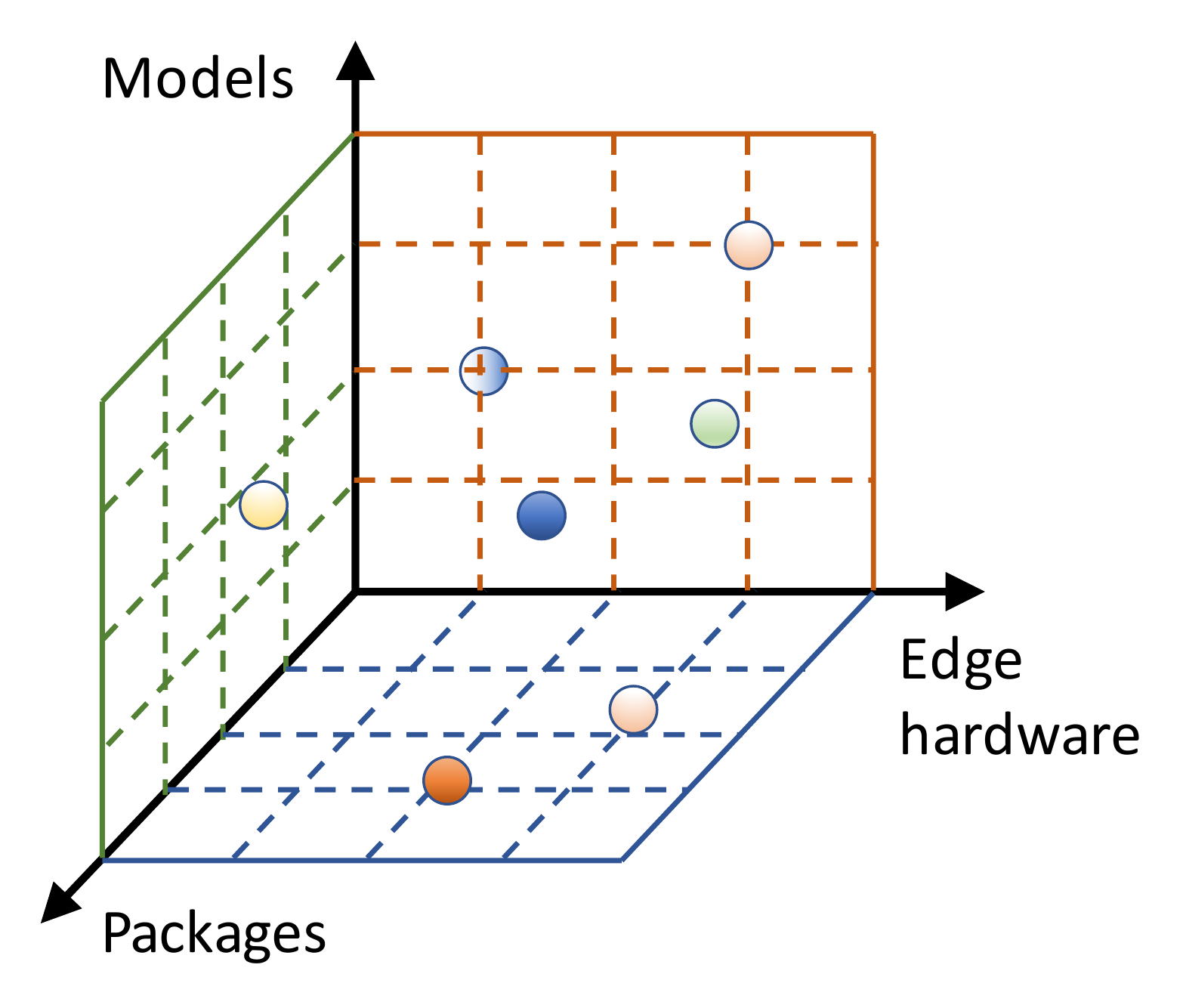}
	\caption{Model selector.}
	\label{fig:Model selector.}
\end{figure}

Currently, neural network based models have started to trickle in. We envision that the biggest problem is not the lack of models, but how to select a matched model for a specific edge based on different \ei capabilities. 
\rev{The \modelselector includes multiple optimized \ai models and a selecting algorithm (SA). The optimized models have been optimized to present a better performance on the \packagemanager based on the techniques which will be discussed in detail in Section IV.A.
Model selecting can be regarded as a multi-dimensional space selection problem. As is shown in Figure \ref{fig:Model selector.}, there are at least three dimensions to choose, e.g., \ai models, machine learning packages, and edge hardware. Taking image classification as an example, more than 10 \ai models (AlexNet, Vgg, ResNet, MobileNet, to name a few), 5 packages (TensorFlow, PyTorch, MXNet, to name a few), and 10 edge hardware platforms (NVIDIA Jetson TX2, Intel Movidius, Mobile Phone, to name a few) need to be considered.}

SA in \modelselector is designed to find the most suitable models for the specific edge platform based on users' requirements. It will first evaluate the \ei capability of the hardware platform based on the four-element tuple \textit{ALEM} $<Accuracy, Latency, Energy, Memory\ footprint>$ and then selecting the most suitable combinations, which is regarded as an optimization problem:

\begin{align}
\begin{split}
\mathop{\arg\min}_{m\in {Models}} \qquad & L \\
s.t. \qquad & A\geq A_{req},  E\leq E_{pro},  M\leq M_{pro}
\end{split}
\end{align}
where $A, L, E, M$ refer to \textit{Accuracy, Latency, Energy, Memory footprint} when running the models on the edge. $A_{req}$ denotes the lowest accuracy that meet the application's requirement. $E_{pro}$ and $M_{pro}$ are the energy and memory footprint that the edge provides. $m$ refers to the selected models and $Models$ refers to all the models.
Equation 1 depicts the desire to minimize $Latency$ while meeting the $Accuracy$, $Energy$ and $Memory footprint$ requirements.
\rev{Meanwhile, if users pay more attention to $Accuracy$, the optimization target will be replaced by maximize $A$ and the constraints are $L$, $E$, and $M$. The same is true of other requirements, i.e. $Energy$ and $Memory footprint$.}
Deep reinforcement learning will be leveraged to find the optimal combination.

\subsection{\libei}

\begin{figure}[!htp]
	\centering
	\includegraphics[width=3.3in]{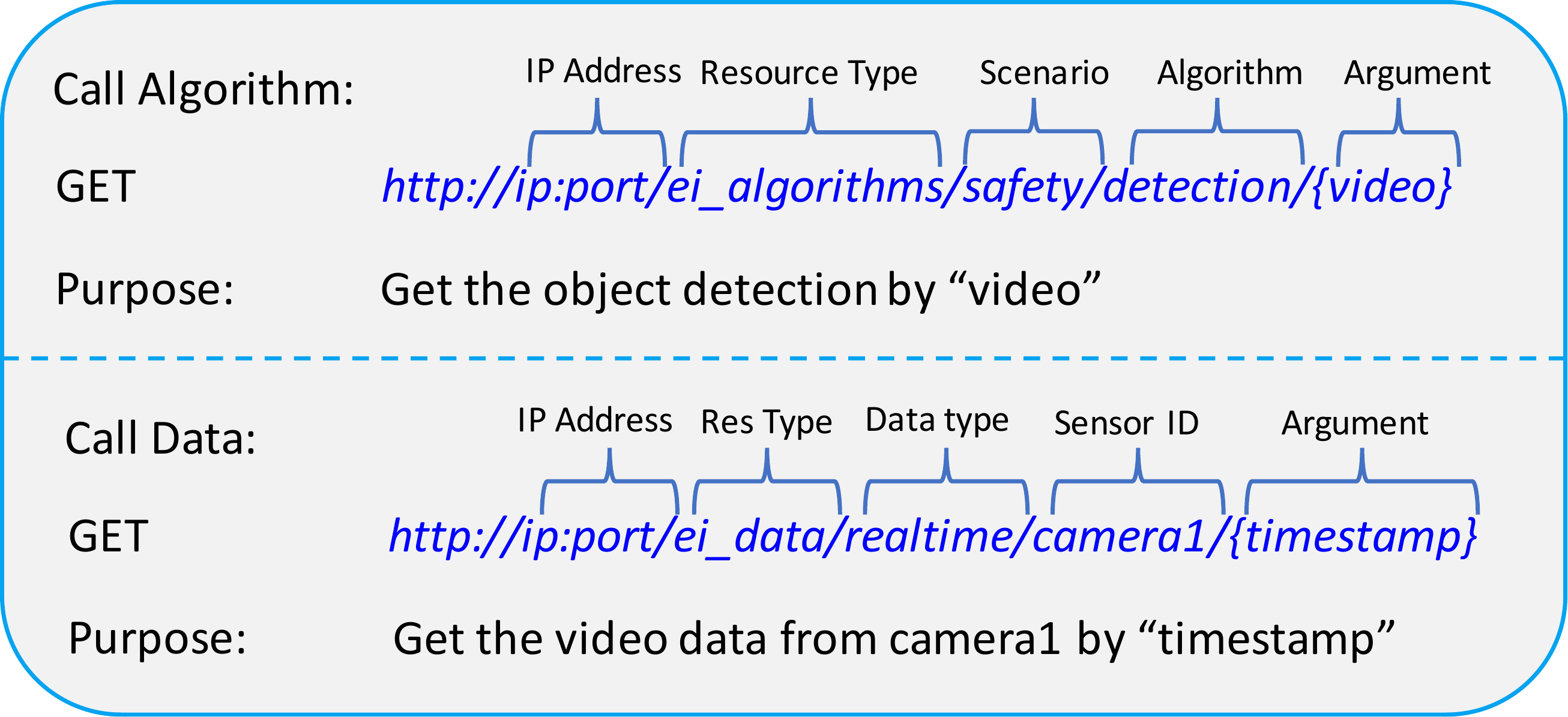}
	\caption{The RESTful API of \textit{libei}}
	\label{fig:libei.}
\end{figure}

\rev{\libei provides a RESTful API which makes it possible to communicate and work together with the cloud, other edges, and IoT devices. Every resource, including the data, computing resource, and models, are represented by a URL whose suffix is the name of the desired resource. As is shown in Figure \ref{fig:libei.}, the RESTful API provided by \libei consists of four fields. The first field is the IP address and port number of the edge. The second field represents the type of recourse, including the algorithm whose suffix is $ei\_algorithms$ and the data whose suffix is $ei\_data$. If users call for the algorithm, the third field indicates the application scenario that \oei supports, including connected vehicles, public safety, smart home, and connected health. The last field is the specific algorithm that the application scenario needs. The argument is the parameter required for algorithm execution. In terms of calling for data APIs, the third field indicates the data's type, including real-time data and historical data and the last field represents the sensor's ID. Developers will get the data over a period of time by the start and end which are provided by the timestamp argument.}

\rev{
\subsection{Summary}
At last, let echo the original example of building an object detection application on the Raspberry Pi to introduce the programming model and summarized the processing flow of \oei. When \oei has been deployed on the Raspberry Pi, the developer is able to visit http://ip:port/ei\_data/realtime/camera1/timestamp=present\_time to get the real-time video frames which could save on the Raspberry Pi. Subsequently, the URI http://ip:port/ei\_algorithms/safety/detection/video=video will be visited to call for the object detection function and response the detection results to the developer.

In terms of the processing flow of \oei, when \libei receives the instruction of object detection, the model selector will choose a most suitable model from the optimized models based on the developer's requirement (the default is accuracy oriented) and the current computing resource of the Raspberry Pi. Subsequently, \packagemanager will call the deep learning package to execute the inference task. If the application is urgent, the real-time machine learning module will be called to guarantee the latency.
}

\section{Key techniques}
\label{sec: Key techniques}
Many techniques from \ai and \ec promote the development of \ei. In this section, we summarize the key techniques and classify them into four aspects: algorithms, packages, running environments and hardware. The implement of \oei will leverage the latest techniques in algorithms and packages. \oei will be installed on the running environments and the performance of the hardware will offer significant references when designing the model selector of \oei. 

\subsection{Algorithms}
Although neural networks are currently widely deployed in academia and industry to conduct nonlinear statistical modeling with state-of-the-art performance on problems that were previously thought to be complex, it is difficult to deploy neural networks on edge devices with limited resources due to their intensive demands on computation and memory. To address this limitation, two main categories of solutions have been recently proposed. One category refers to the deep model compression method, which aims to aid the application of current advanced networks in devices. The other is the \ei algorithm, which refers to the efficient machine learning algorithms that we developed to run on the resource-constrained edges directly.

\subsubsection{Compression}
Compression techniques are roughly categorized into three groups: Parameter sharing and pruning methods, low-rank approximation methods, and knowledge transfer methods \cite{cheng2017survey, han2015deep}.

Parameter sharing and pruning method control the capacity and storage cost by reducing the number of parameters which are not sensitive to the performance. 
This method only needs to save the values of these representatives and the indexes of these parameters. Courbariaux \etal \cite{courbariaux2015training} proposed a binary neural network to quantify the weights. More specifically, it refers to restricting the value of the network weight by setting it to -1 or 1, and it simplifies the design of hardware that is dedicated to deep learning. Gong \etal \cite{gong2014compressing} employed the k-means clustering algorithm to quantize the weights of fully connected layers, which could achieve up to 24 times the compression of the network with only 1\% loss of classification accuracy for the CNN network in the ImageNet challenge. Chen \etal  \cite{chen2015compressing}presented a HashedNets weight sharing architecture that groups connection weights into hash buckets randomly by using a low-cost hash function, where all connections of each hash bucket have the same value. The values of the parameters are adjusted using the standard backpropagation method \cite{werbos1990backpropagation} during training. 
Han \etal \cite{han2015learning} pruned redundant connections using a three-step method: First, the network learns which connections are important, and then they prune the unimportant connections. Finally, they retrain the network to fine tune the weights for the remaining connections.


Low-rank approximation refers to reconstructing the dense matrix to estimate the representative parameters. Denton \etal \cite{denton2014exploiting} use singular value decomposition to reconstruct the weight of all connected layers, and they triple the speedups of convolutional layers on both CPU and GPU, and the loss of precision is controlled within 1\%. Denil \etal \cite{denil2013predicting} employ low-rank approximation to compress the weights of different layers and reduce the number of dynamic parameters. Sainath \cite{sainath2013low} uses a low-rank matrix factorization on the final weight layer of a DNN for acoustic modeling.

Knowledge transfer is also called teacher-student training. The idea of knowledge transfer is to adopt a teacher-student strategy and use a pre-trained network to train a compact network for the same task\cite{sau2016deep}. It was first proposed by Caruana \etal \cite{bucilu2006model}. They used a compressed network of trained network models to mark some unlabeled simulation data and reproduced the output of the original larger network. The work in \cite{balan2015bayesian} trained a parametric student model to estimate a Monte Carlo teacher model. Ping \etal \cite{luo2016face} use the neurons in the hidden layer to generate more compact models and preserve as much of the label information as possible. Based on the idea of function-preserving transformations, the work in \cite{chen2015net2net} instantaneously transfers the knowledge from a teacher network to each new deeper or wider network. 


\begin{table}[h]
\caption{Typical approaches for deep compression.} 
\label{tab:deepcompression}
\scalebox{0.83}{
\begin{tabular}{|l|l|l|l|}
\hline
\multicolumn{1}{|c|}{\textbf{Method}} & \multicolumn{1}{c|}{\textbf{Description}} & \multicolumn{1}{c|}{\textbf{Advantages}} & \multicolumn{1}{c|}{\textbf{Disadvantages}} \\ \hline
\begin{tabular}[c]{@{}l@{}}Parameter\\ sharing\\ and pruning\end{tabular} & \begin{tabular}[c]{@{}l@{}}Reducing uninfor-\\ mative parameters\\ that are not sen-\\ sitive to the perfor-\\ mance\end{tabular} & \begin{tabular}[c]{@{}l@{}}Robust to various\\ settings, su-\\pport training\\ from scratch and\\ pre-trained model\end{tabular} & \begin{tabular}[c]{@{}l@{}}Pruning requires man-\\ ual setup of sensitiv-\\ ity for layers, which\\ demands fine-tuning\\ of the parameters and\\ may be cumbersome\\ for some applications.\end{tabular} \\ \hline
\begin{tabular}[c]{@{}l@{}}Low-rank\\ factorization\end{tabular} & \begin{tabular}[c]{@{}l@{}}Using the matrix\\ decomposition\\ method to figure\\ out the represe-\\ ntative parame-\\ ters\end{tabular} & \begin{tabular}[c]{@{}l@{}}Straightforward for \\ model compression,\\ standardized pipe-\\ line, support\\ pre-trained\\ models and training\\ from scratch\end{tabular} & \begin{tabular}[c]{@{}l@{}}The implementation\\ involves the decom-\\ position operation, \\ which is computati-\\ onally expensive\end{tabular} \\ \hline
\begin{tabular}[c]{@{}l@{}}Knowledge\\ transfer\end{tabular} & \begin{tabular}[c]{@{}l@{}}Using a pre-trained\\ neural network to\\ train a compact\\ network on the same\\ task\end{tabular} & \begin{tabular}[c]{@{}l@{}}Make deeper\\ models thinner,\\ significantly\\ reduce the \\ computational\\ cost\end{tabular} & \begin{tabular}[c]{@{}l@{}}Only be applied\\ to the classification\\ tasks with softmax\\ loss functions, netw-\\ ork structure only\\ support training from\\ scratch\end{tabular} \\ \hline
\end{tabular}}
\end{table}
Table~\ref{tab:deepcompression} concludes the above three typical compression technologies, and describes the advantages and disadvantages of each technology.

\subsubsection{\ei algorithm}
In this paper, we define \ei algorithms as the those designed for the resource-constrained edges directly. Google Inc. \cite{howard2017mobilenets} presented efficient CNN for mobile vision
applications, called MobileNets. The two hyper-parameters that Google introduced allow the model builder to choose the right sized model for the specific application. It not only focuses on optimizing for latency but also builds small networks. MobileNets are generated mainly from depth-wise separable convolutions, which were first introduced in the work of \cite{sifre2014rigid} and subsequently employed in Inception models \cite{ioffe2015batch}. Flattened networks \cite{jin2014flattened} are designed for fast feedforward
execution. They consist of a consecutive sequence of one-dimensional filters that span every direction of three-dimensional space to achieve comparable performance as conventional convolutional networks \cite{wang2017factorized}. Another small network is the Xception network \cite{chollet2017xception}; Chollet \etal proposes the dubbed Xception architecture inspired by Inception V3, where Inception modules have been replaced with depthwise separable convolutions. It shows that the architecture slightly outperforms Inception V3 on the ImageNet data set. Subsequently, Iandola \etal \cite{iandola2016squeezenet} developed Squeezenet, a small CNN architecture. It achieves AlexNet-level \cite{krizhevsky2012imagenet} accuracy with 50 times fewer parameters on ImageNet data set (510 times smaller than AlexNet).

In 2017, Microsoft Research India proposed Bonsai \cite{kumar2017resource} and ProtoNN\cite{gupta2017protonn}. Then, they developed EMI-RNN \cite{dennis2018multiple} and FastGRNN \cite{kusupati2018fastgrnn} in 2018. Bonsai \cite{kumar2017resource} refers to a tree-based algorithm used for efficient prediction on IoT devices. More specifically, it is designed for supervised learning tasks such as regression, ranking, and multi-class classification, etc. ProtoNN \cite{gupta2017protonn} is inspired by k-Nearest Neighbor (KNN) and could be deployed on the edges with limited storage and computational power (e.g., an Arduino UNO with 2kB RAM) to achieve excellent prediction performance. EMI-RNN \cite{dennis2018multiple} requires 72 times less computation than standard Long Short-term Memory Networks (LSTM) \cite{hochreiter1997long} and improving accuracy by 1\%. 


\emph{Open Problems}: Here are three main open problems that need to be addressed to employ \ei algorithms on edges. First, to reduce the size of algorithms, many techniques have been proposed to reduce the number of connections and parameters in neural network models. However, the pruning process usually affects algorithm accuracy. Hence, how to reduce the model size while guaranteeing high accuracy is a research direction in the \ei area. Second, collaboration between edges calls for an algorithm that runs in a distributed manner on multiple edges. It is a challenge to research how to split an algorithm based on the computing resources of the edges. Third, how to achieve collaborative learning on the cloud and edges is also a research direction.

\subsection{Packages}

In order to execute AI algorithms efficiently, many deep learning packages are specifically designed to meet the computing paradigm of AI algorithms, such as TensorFlow, Caffe, MXNet, and PyTorch. However, these packages are focused on the cloud and not suitable for the edge. 
On the cloud, packages use a large-scale dataset to train deep learning models. One of the main tasks of packages is to learn a number of weights in each layer of a model. They are deployed on the high-performance platforms, such as GPU, CPU, FPGA, and ASIC (TPU \cite{jouppi2017datacenter}) clusters. 
On the edges, due to limited resources, packages do not train models in most cases. 
They carry on inference tasks by leveraging the models which have been trained in the cloud. The input is small-scale real-time data and the packages are installed on the heterogeneous edges, such as edge server, mobile phone, Raspberry Pi, laptop, etc.


To support processing data and executing AI algorithms on the edges, several edge-based deep learning packages 
have been released by some top-leading tech-giants.
Compared with cloud versions, these frameworks require significantly fewer resources, but behave almost the same in terms of inference. 
TensorFlow Lite \cite{TensorFlow-Lite} is TensorFlow’s lightweight solution which is designed for mobile and edge devices. It leverages many optimization techniques, including optimizing the kernels for mobile apps, pre-fused activations, and quantized kernels to reduce the latency.  Apple published CoreML \cite{CoreML}, a deep learning package optimized for on-device performance to minimizes memory footprint and power consumption. Users are allowed to integrate the trained machine learning model into Apple products, such as Siri, Camera, and QuickType. 
Facebook developed QNNPACK (Quantized Neural Networks PACKage) \cite{qunnpack}, which is a mobile-optimized library for high-performance neural network inference. It provides an implementation of common neural network operators on quantized 8-bit tensors.

In the meantime, cloud-based packages are also starting to support edge devices, such as MXNet \cite{chen2015mxnet} and TensorRT \cite{TensorRT}. MXNet is a flexible and efficient library for deep learning. It is designed to support multiple platforms (either cloud platforms or edge ones) and execute training and inference tasks. TensorRT is a platform for high-performance deep learning inference, not training and will be deployed on the cloud and edge platforms. Several techniques, including weight and activation precision calibration, layer and tensor fusion, kernel auto-tuning, and multi-stream execution are used to accelerate the inference process.


Zhang \etal made a comprehensive performance comparison of several state-of-the-art deep learning frameworks on the edges and evaluated the latency, memory footprint, and energy of these frameworks with two popular deep learning models on different edge devices \cite{zhang18pcamp}. They found that no framework could achieve the best performance in all dimensions, which indicated that there was a large space to improve the performance of AI frameworks on the edge. It is very important and urgent to develop a lightweight, efficient and highly-scalable framework to support AI algorithms on edges.


\emph{Open Problems}: There are several open problems that need to be addressed to be able to build data processing frameworks on the edge. First, to execute real-time tasks on the edge, many packages sacrifice memory to reduce latency. However, memory on the edge is also limited. Thus, how to tradeoff the latency and memory? Second, having access to personalized, training on the edge is ideal while the training process usually requires huge computing resources. Therefore, how to implement a local training process with limited computing power? Last, with the support of \oei, the edge will need to handle multiple tasks which raises the problem of how to execute multiple tasks on a package in the meantime.

\subsection{Running environments}
\rev{The most typical workloads from \ei are model inference and collaborative model training, so the \ei running environments should be capable of handling deep learning packages,
allocating computation resources and migrating computation loads. Meanwhile, they should be lightweight enough and can be deployed on heterogeneous hardware platforms. 
Taking the above characteristic into account, some studies like TinyOS, ROS, and OpenVDAP are recognized as potential systems to support \ei.


TinyOS\cite{levis2005tinyos} is an application based operating system for sensor networks. The biggest challenge that TinyOS has solved is to handle concurrency intensive operations with small physical size and low power consumption\cite{hill2000system}. TinyOS takes an event-driven design which is composed of a tiny scheduler and a components graph.
The event-driven design makes TinyOS achieve great success in sensor networks. However, enabling effective computation migration is still a big challenge for TinyOS.}

Robot Operating System(ROS)\cite{quigley2009ros} is recognized as the typical representative of next the generation of mobile operating systems to cope with the Internet of Things. 
In ROS, the process that performs computations is called a node. For each service, the program or features are divided into several small pieces and distributed on several nodes, and the ROS topic is defined to share messages between ROS nodes. The communication-based design of ROS gives it high reusability for robotics software development. Meanwhile, the active community and formation of the ecosystem put ROS in a good position to be widely deployed for edge devices. However, as ROS is not fundamentally designed for resource allocation and computation migration, there are still challenges in deploying \ei service directly on ROS.

OpenVDAP \cite{zhang2018openvdap} is an edge based data analysis platform for Connected and Autonomous Vehicles(CAVs). OpenVDAP is a full stack platform which contains Driving Data Integrator($DDI$), Vehicle Computing Units($VCU$), edge-based vehicle operating system($EdgeOS_v$), and libraries for vehicular data analysis($libvdap$). Inside OpenVDAP, $VCU$ supports \ei by allocating hardware resources according to an application, and $libvdap$ supports \ei by providing multi-versions of models to accelerate the model inference.

\emph{Open Problems}: There is a crucial open problem that needs to be addressed: how to design a lightweight edge operating system with high availability. For the scenario with dynamic changes in topology and high uncertainty in wireless communication, the edge operating system calls for high availability related to the consistency, resource management, computation migration, and failure avoidance. Meanwhile, the edge operating system should be light enough to be implemented on the computing resource-constraint edge. 

\subsection{Hardware}
\rev{Recently, heterogeneous hardware platforms present the potential to accelerate specific deep learning algorithms while reducing the processing time and energy consumption \cite{nurvitadhi2016acceleratingr,nurvitadhi2016acceleratingb}. For the hardware on \ei, various heterogeneous hardware are developed for particular \ei application scenario to address the resource limitation problem in the edge.}

ShiDianNao \cite{du2015shidiannao} first proposed that the artificial intelligence processor should be deployed next to the camera sensors. The processor accesses the image data directly from the sensor instead of DRAM, which reduces the power consumption of sensor data loading and storing. 
ShiDianNao is 60 times more energy efficient and 30 times faster than the previous state-of-the-art AI hardware, so it will be suitable for the \ei applications related to computer vision. EIE\cite{han2016eie} is an efficient hardware design for compressed DNN inference. It leverages multiple methods to improve energy efficiency, such as exploiting DNN sparsity and sharing DNN weights, so it is deployed on mobile devices to process some embedded \ei applications. In industry, many leaders have published some dedicated hardware modules to accelerate \ei applications; for example, IBM TrueNorth\cite{modha2017introducing} and Intel Loihi\cite{davies2018loihi} are both the neuromorphic processors.

In addition to ASICs, several studies have deployed FPGAs or GPUs for \ei application scenarios, such as speech recognition.
ESE\cite{han2017ese} used FPGAs to accelerate the LSTM model on mobile devices, which adopted the load-balance-aware pruning method to ensure high hardware utilization and the partitioned compressed LSTM model on multiple PEs to process LSTM data flow in parallel. The implementation of ESE on Xilinx FPGA achieved higher energy efficiency compared with the CPU and GPU. Biookaghazadeh \etal used a specific \ei workload to evaluate FPGA and GPU performance on the edge devices. They compared some metrics like data throughput and energy efficiency between the FPGA and GPU. The evaluation results showed that the FPGA is more suitable for \ei application scenarios\cite{biookaghazadeh2018fpgas}. In industry, NVIDIA published the Jetson AGX Xavier module\cite{nvidia2019jetson}, which is equipped with a 512-core Volta GPU and an 8-core ARM 64-bit CPU. It supports the CUDA and TensorRT libraries to accelerate \ei applications in several scenarios, such as robot systems and autonomous vehicles.

\emph{Open Problems}: There are several open problems that need to be addressed to design a hardware system for \ei scenarios. First, novel hardware designed for \ei has improved the processing speed and energy efficiency; hence, the question remains whether there is any relationship between the processing speed and power. For example, if the processing power is limited, we need to know how to calculate the maximum speed that the hardware reaches. Second, the \ei platform may be equipped with multiple types of heterogeneous computing hardware, so managing the hardware resource and scheduling the \ei application among the types of hardware to ensure high resource utilization are important questions. Third, we need to be able to evaluate how suitable the hardware system is for each specific \ei application.
\section{Typical Application Scenarios}
\label{sec: Potential applications}

With the development of \ei techniques, many novel applications are quickly emerging, such as live video analytics for public safety, connected and autonomous driving, smart homes, and smart and connected health care services. As shown in Figure \ref{fig:The overview of OpenEI.}, \oei provides RESTful API to support these \ai scenarios.
This section will illustrate the typical application scenarios and discuss how to leverage \oei to support these applications.

\subsection{Video Analytics in Public Safety}

Video Analytics in Public Safety(VAPS) is one of the most successful applications on edge computing since it has the high real-time requirements and unavoidable communication overhead. \oei is used to deploy on cameras or edge severs to support VAPS and provide an API for the user. Third-party developers execute the widely used algorithms on public safety scenarios by calling \textit{http://ip:port/ei\_algorithms/safety/} plus the name of the algorithms.
The applications of video analysis for public safety that \oei supports are divided into the following two aspects.

The first aspect is from the algorithm perspective, which is aimed at designing a lightweight model to support \ei. 
The strength of edge devices is that the data is stored locally so there is no communication delay. However, one drawback is that edge devices are not powerful enough to implement large neural networks; the other is that the vibration in a video frame makes it more difficult to analyze. 
For example, criminal scene auto detection is a typical application of VAPS. The challenges are created by real-time requirements and the mobility of the criminal. Hence, the model should do preprocessing on each frame to evict the influence of mobility. 
In addition to criminal scene auto detection, for some applications like High-Definition Map generation, masking some private information like people's face is also a potential VAPS application. 
The objective is to enable the edge server to mask the private information before uploading the data. Video frame preprocessing at the edge supports \ei by accelerating the model training and inference process.

The second aspect is from the system perspective, which enables edge devices like smartphones and body cameras to run machine learning models for VAPS applications. In \cite{zhang2016demo}, an edge based real-time video analysis system for public safety is proposed to distribute the computing workload in both the edge node and the cloud in an optimized way. A3\cite{zhang2018distributed} is an edge based amber alert application which support distributed collaborative execution on the edge. SafeShareRide\cite{8567654} is an edge based detection platform which enables a smartphone to conduct real-time detection including video analysis for both passengers and drivers in ridesharing services. Moreover, a reference architecture which enables the edge to support VAPS applications is also crucial for \ei. In~\cite{liu2019autovaps}, Liu~\etal proposed a reference architecture to deploy VAPS applications on police vehicles. \ei will be supported through efficient data management and loading.

\subsection{Connected and Autonomous Vehicles} 

Today, a vehicle is not just a mechanical device but is gradually becoming an intelligent, connected, and autonomous system. We call these advanced vehicles connected and autonomous vehicles (CAVs). 
CAVs are significant application scenarios for \ei and many applications on CAVs are tightly integrated into \ei algorithms, such as localization, object tracking, perception, and decision making. \oei also provides API for the CAVs scenarios to execute the \ai algorithm on a vehicle. The input is the video data collected by on-board cameras.

The autonomous driving scenario has conducted many classic computer vision and deep learning algorithms\cite{mur2017orb,liu2016ssd}. Since the algorithms will be deployed on the vehicle, which is a resource-constrained and real-time \ec system, the algorithm should consider not only precision but also latency, as the end-to-end deep learning algorithm YOLOv3\cite{redmon2018yolov3}. To evaluate the performance of algorithms in the autonomous driving scenario, Geiger \etal published KITTI benchmark datasets\cite{geiger2012we}, which provide a large quantity of camera and LiDAR data for various autonomous driving applications.

Lin \etal explored the hardware computing platform design of autonomous vehicles\cite{lin2018architectural}. They chose three core applications on autonomous vehicles, which are localization, object detection, and object tracking, to run on heterogeneous hardware platform: GPUs, FPGAs, and ASICs. According to the latency and energy results, they provided the design principle of the end-to-end autonomous driving platform. From the industry, NVIDIA published the DRIVE PX2 platform for autonomous vehicles\cite{nvidia2019nvidia}. To evaluate the performance of the computing platform designed for CAVs, Wang \etal first proposed CAVBench\cite{wang2018cavbench}, which takes six diverse on-vehicle applications as evaluation workloads and provides the matching factor between the workload and the computing platform.

In the real world, we still need a software framework to deploy \ei algorithms on the computing platform of connected and autonomous vehicle. OpenVDAP\cite{zhang2018openvdap}, Autoware\cite{kato2015open}, and Baidu Apollo\cite{baidu2019apollo} are open-source software frameworks for autonomous driving, which provide interfaces for developers to build and customize  autonomous driving vehicles.

\subsection{Smart Homes}
Smart homes have become popular and affordable with the development of~\ec and~\ai technologies. By leveraging different types of IoT devices (e.g., illuminate devices, temperature and humidity sensors, surveillance system, etc.), it is feasible to keep track of the internal state of a home and ensure its safety, comfort, and convenience under the guarantee of~\ei. The benefit of involving~\ei in a smart home is twofold. First, home privacy will be protected since most of the computing resources are confined to the home internal gateway and sensitive family data is prohibited from the outflow. Second, the user experience will be improved because the capability of intelligent edge devices facilitates the installation, maintenance, and operation of a smart home with less labor demanded. 
As an important \ei scenario, \oei provides APIs to call the \ai algorithms related to the smart home. \textit{http://ip:port/ei\_algorithms/home/power\_monitor} is used to call to execute the power monitoring algorithms on the edge.

Intelligence in the home has been developed to some extent, and related products are available on the market. As one of the most intelligent devices in the smart home ecosystem, smart speaker such as Amazon Echo~\cite{amazonEcho}, Google Home~\cite{googleHome} are quite promising models that involve in~\ei. They accept the user's instructions and respond accordingly by interacting with a third party service or household appliances. From timing to turning off lights, from memo to shopping, their intelligence enhances people's quality of life significantly. Despite this, the utility of the edge is not well reflected and utilized in this technology. Relying on cloud cognitive services, smart speakers need to upload data to the cloud and use deep neural networks for natural language understanding and processing, which becomes a hidden danger of family data privacy leakage and increases the burden of unnecessary network transmission. \ei is the principal way to solve these problems. 


Considering the privacy of the home environment and the accessibility of smart home devices, it is completely feasible and cost-effective to offload intelligent functions from the cloud to the edge, and there have been some studies demonstrating \ei capabilities. Wang \etal found that a smart home will benefit from~\ei to achieve energy efficiency~\cite{wang2017poweranalyzer}. Zhang \etal developed IEHouse, a non-intrusive status recognition system, for household appliance~\cite{zhang2017iehouse} with the assistance of deep neural networks. Zhang~\etal proposed a CNN model running on edge devices in a smart home to recognize activity with promising results~\cite{zhang2018enabling}. In addition to indoor activity detection, surveillance systems play an important role in protecting the home security both indoor and outside. Because the video stream occupies a considerable storage space and transmission bandwidth, it is almost impossible to upload every frame recorded from a surveillance system to the cloud for further processing, especially for high resolution videos~\cite{abdallah2017lessons}. \ei enables a surveillance device to have certain image processing capabilities, such as object recognition and activity detection, to extract valid information from redundant videos to save unnecessary computing and storage space. 

Home entertainment systems also benefit from~\ei to provide a better user experience. 
A home audio and video system is one typical example. With~\ei involved, the system handles the user's personalized recommendation service by itself, without uploading any privacy data about the user's preferences to the cloud, so that the user has a smoother and safer entertainment experience.
Meanwhile, with the maturity of Augmented Reality and Virtual Reality technology, users are able to have a better game immersive experience. MUVR is proposed in this scenario to boost the  multi-user gaming experience with the edge caching mechanism~\cite{li2018muvr}. Motion sensing games are a typical example. \ei gives it the capability to detect action and behavior without equipping users with a control bar or body sense camera. 


\subsection{Smart and Connected Health}
Health and biomedicine are entering a data-driven epoch~\cite{martin2014big,andreu2015big}. On the one hand, the development of medical instruments indicates health status with data. On the other hand, the popularization of health awareness has led citizens to track their physical condition with smart edge devices.  
Similar to the other three scenarios above, \oei also provides the API to call for the related algorithms. The~\ei of smart health is quite promising and is created from the following aspects.

First is pre-hospital emergency medicine, where the emergent patient is been cared for before reaching the hospital, or during emergency transfer between hospitals, emergency medical service (EMS) systems are provided in the form of basic life support (BLS) and advanced life support (ALS). Current EMS systems focus on responsiveness and transportation time, while the health care solutions are traditional and less efficient, some of which have been used since the 1990s. Although ALS is equipped with higher level care, the number of ALS units is highly constrained because of limited budgets~\cite{wu2017strems}. In addition, the data transmission is greatly affected by the moving scenario and the extreme weather in the cloud computing. Considering the limitation of the status quo,~\ei is an alternative way to enhance EMS quality in terms of responsiveness and efficiency by building a bidirectional real time communication channel between the ambulance and the hospital, which has intelligent features like natural language processing, and image processing. 

Second is smart wearable sensors. Most of the current technologies for smart wearable sensors are based on cloud computing because of the limitations of computing resources and capabilities. That is, wearable sensors are more like a data collector than a data analyst. Fortunately, \ei research in this field is emerging. 
Rincon~\etal deployed an artificial neural network on wearable sensors to detect emotion~\cite{rincon2016using}. 
With the promising development of~\ec, there will be more light-weight intelligent algorithms running on smart wearable devices to monitor, analyze, and predict health data in a timely manner, which it will ease the pressure on caregivers and doctors, and let users have better knowledge of their physical condition. 

Third is the preservation and sharing of medical data. The US government has promoted the sharing of personal health information since 2009~\cite{hitech}, but it turns out that although doctors' usage of electronic medical records has improved, interoperability is missing, and the large amount of medical data gathered together does not produce real value. \ei improves the status quo by training the sharing system to mark and recognize medical images, promote communication, and improve treatment efficiency.

\section{Conclusion}
\label{sec: Conclusion}

With the development of \ai and \ec, \ei emerges since it has the potential to reduced bandwidth and cost while maintaining the quality of service compared to processing on the cloud. 
In this paper, we define \ei as a capability that enables edges to execute artificial intelligence algorithms. 
To support \ei, several techniques are being developed, including algorithms, deep learning packages, running environments and hardware. This paper discussed the challenges that these techniques brings and illustrated four killer applications in the \ei area.

To address the challenges for data analysis of \ei, computing power limitation, data sharing and collaborating, and the mismatch between the edge platform and AI algorithms, we presented an Open Framework for Edge Intelligence (\oei) which is a lightweight software platform to equip the edge with intelligent processing and data sharing capability.
We hope that \oei will be used as a model for prototyping in \ei. We also hope that this paper provides helpful information to researchers and practitioners from various disciplines when designing new
technologies and building new applications for \ei.

\newpage
\bibliographystyle{IEEEtran}
\bibliography{EI}

\end{document}